\documentclass[letterpaper]{article} 
\usepackage{aaai2026}  
\usepackage{times}  
\usepackage{helvet}  
\usepackage{courier}  
\usepackage[hyphens]{url}  
\usepackage{graphicx} 
\usepackage{natbib}  
\usepackage{caption} 
\usepackage{algorithm}
\usepackage{algorithmic}

\usepackage{booktabs}  
\usepackage{array}     
\usepackage{multirow}
\usepackage{colortbl}
\usepackage{makecell}
\usepackage{arydshln}  

\usepackage{amsmath}
\usepackage{amsfonts} 

\usepackage{bbold}

\definecolor{lightgreen}{RGB}{198, 239, 206}
\definecolor{lightcoral}{RGB}{255, 199, 206}
\definecolor{lightgray}{RGB}{245,245,245}

\usepackage{graphicx}
\usepackage{subcaption}

\usepackage{tcolorbox}
\usepackage{newfloat}
\usepackage{listings}
\DeclareCaptionStyle{ruled}{labelfont=normalfont,labelsep=colon,strut=off} 
\floatstyle{ruled}
\newfloat{listing}{tb}{lst}{}
\floatname{listing}{Listing}
\title{SOI is the Root of All Evil: Quantifying and Breaking Similar Object Interference in Single Object Tracking}

\author {
    Yipei Wang\textsuperscript{\rm 1},
    Shiyu Hu\textsuperscript{\rm 2},
    Shukun Jia\textsuperscript{\rm 1},
    Panxi Xu\textsuperscript{\rm 3},
    Hongfei Ma\textsuperscript{\rm 3},
    Yiping Ma\textsuperscript{\rm 4}, 
    Jing Zhang\textsuperscript{\rm 5}, \\
    Xiaobo Lu\textsuperscript{\rm 1}\thanks{Corresponding author},
    Xin Zhao\textsuperscript{\rm 3}
}

\affiliations {
    \textsuperscript{\rm 1}Southeast University, China
    \textsuperscript{\rm 2}Nanyang Technological University, Singapore
    \textsuperscript{\rm 3}University of Science and Technology Beijing, China
    \textsuperscript{\rm 4}East China Normal University, China
    \textsuperscript{\rm 5}Chinese Academy of Sciences, China\\
    wyppyw@seu.edu.cn, 
    shiyu.hu@ntu.edu.sg, 
    jia\_shukun@seu.edu.cn, 
    panxixu@ustb.edu.cn, 
    hongfeima@ustb.edu.cn, \\
    mayiping98@163.com, 
    jing\_zhang@ia.ac.cn,
    xblu2013@126.com, 
    zhaoxin@ustb.edu.cn
}

\usepackage{bibentry}

\begin{document}

\maketitle

\begin{abstract}

In this paper, we present the first systematic investigation and quantification of Similar Object Interference (SOI), a long-overlooked yet critical bottleneck in Single Object Tracking (SOT). Through controlled Online Interference Masking (OIM) experiments, we quantitatively demonstrate that eliminating interference sources leads to substantial performance improvements (AUC gains up to 4.35) across all SOTA trackers, directly validating SOI as a primary constraint for robust tracking and highlighting the feasibility of external cognitive guidance. Building upon these insights, we adopt natural language as a practical form of external guidance, and construct \textbf{SOIBench}—the first semantic cognitive guidance benchmark specifically targeting SOI challenges. It automatically mines SOI frames through multi-tracker collective judgment and introduces a multi-level annotation protocol to generate precise semantic guidance texts. Systematic evaluation on SOIBench reveals a striking finding: existing vision-language tracking (VLT) methods fail to effectively exploit semantic cognitive guidance, achieving only marginal improvements or even performance degradation (AUC changes of -0.26 to +0.71). In contrast, we propose a novel paradigm employing large-scale vision-language models (VLM) as external cognitive engines that can be seamlessly integrated into arbitrary RGB trackers. This approach demonstrates substantial improvements under semantic cognitive guidance (AUC gains up to 0.93), representing a significant advancement over existing VLT methods. We hope SOIBench will serve as a standardized evaluation platform to advance semantic cognitive tracking research and contribute new insights to the tracking research community.

\end{abstract}


\section{Introduction}

\begin{figure}[!t]
\centering
\includegraphics[width=0.5\textwidth]{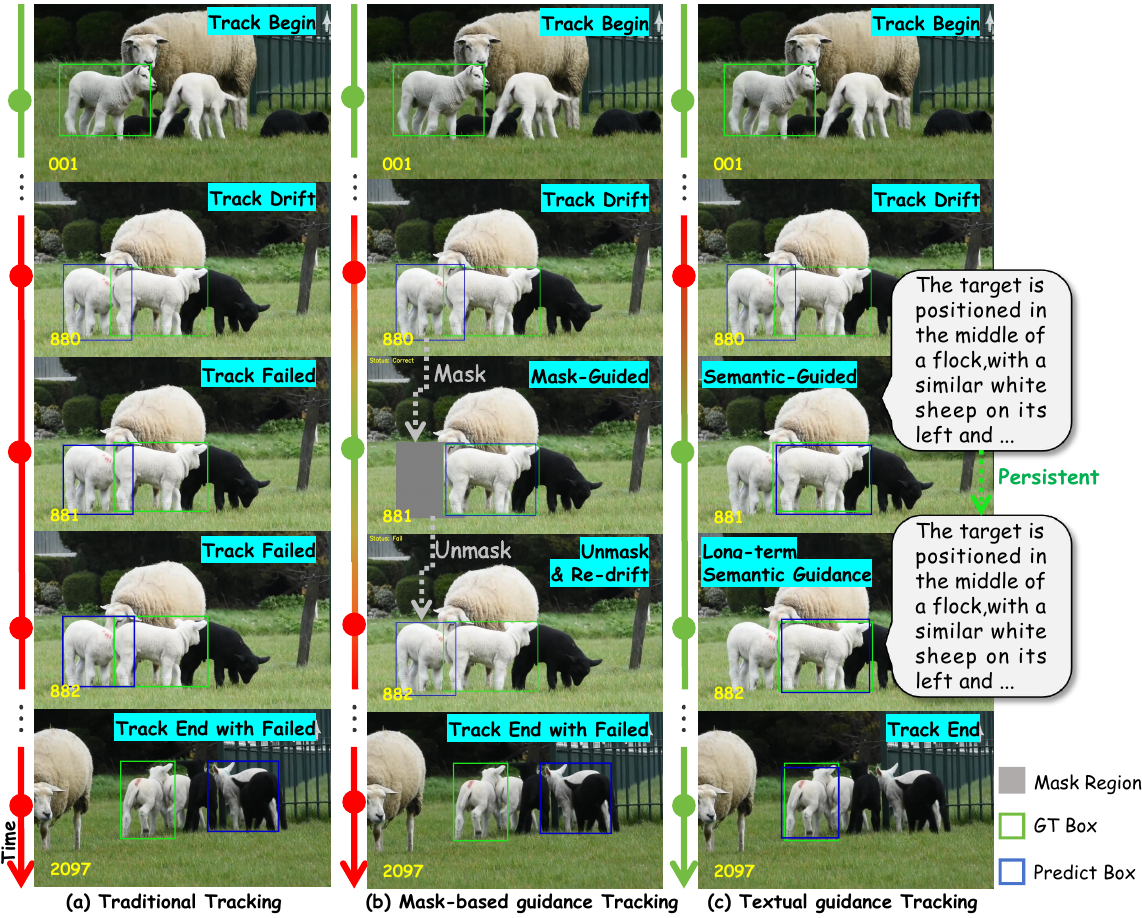}
\caption{SOI as the root of tracking failures: (a) Traditional tracking fails under SOI; (b) Ideal mask guidance proves SOI's impact; (c) Practical semantic guidance achieves cognitive breakthrough.}
\label{fig:soi_mask_vis}
\end{figure}

Single Object Tracking (SOT) is a sequential decision-making task that requires continuous identification and tracking of a designated target \cite{Huang2021got10k}. Among the complex spatio-temporal challenges it faces, Similar Object Interference (SOI) is particularly critical. SOI specifically refers to scenarios where non-target objects closely resemble the target's current appearance, creating ambiguous visual contexts that force trackers to make continuous fine-grained discriminative decisions. As shown in Fig.~\ref{fig:soi_mask_vis} (a), when distractors appear or target features degrade due to occlusion, fast motion, and other factors, existing methods struggle to maintain stable decision-making and suffer persistent tracking drift. 

Despite SOI's widespread existence in critical applications such as surveillance and animal behavior analysis \cite{hu2024multi}, prior work has primarily focused on overall performance improvements without isolating and quantifying SOI's specific contribution to tracking failures. This analytical gap limits our understanding of trackers' essential limitations and hinders targeted solution development.

To systematically quantify the impact of SOI, we design controlled Online Interference Masking (OIM) experiments. As shown in Fig.~\ref{fig:soi_mask_vis} (b), when tracking drift is detected, we eliminate SOI by precisely masking interference sources, implementing a form of direct and strong visual external guidance. The results are striking: all tested state-of-the-art trackers achieve substantial performance improvements (AUC gains up to 4.35), directly validating SOI as a primary constraining factor for robust tracking. These performance gains also demonstrate the feasibility of external guidance for enhancing tracker robustness. However, trackers immediately re-drift once mask-based guidance is removed, revealing a fundamental limitation of existing tracking methods: their reliance on shallow appearance matching rather than genuine semantic understanding of target identity.

The effectiveness yet impracticality of mask-based guidance due to ground-truth (GT) requirements inspired us to think: \textbf{\textit{how to provide external guidance that is both effective and practical?}} From cognitive science, humans invoke multi-level semantic knowledge when distinguishing similar objects—from spatial relationships and appearance attributes to motion states and contextual cues. This inspires us to formalize human semantic reasoning as textual descriptions that serve as external cognitive guidance. Compared to visual masks, natural language offers unique advantages: hierarchical expressiveness, temporal persistence, and deployment flexibility. As shown in Fig.~\ref{fig:soi_mask_vis}(c), we transform infeasible mask guidance into practical semantic textual guidance through target-centered descriptions.

To establish a standardized platform for researching semantic guidance in SOI scenarios, we propose \textbf{SOIBench}—the first comprehensive benchmark specifically designed for SOI challenges with integrated semantic guidance. SOIBench employs multi-tracker collective judgment to automatically mine challenging SOI scenarios, eliminating subjective bias in manual annotation: frames are marked as SOI when most trackers produce high-confidence multi-peak responses. This approach ensures the identified frames correspond to current trackers' actual cognitive limits, creating truly challenging evaluation environments. Furthermore, SOIBench introduces a multi-level textual annotation protocol inspired by cognitive linguistics: positional context (L1), static appearance attributes (L2), dynamic state description (L3), and discriminative cues (L4), providing hierarchical semantic guidance for SOI frames.

Using SOIBench, our systematic evaluation of existing vision-language tracking (VLT) methods reveals a striking limitation: despite access to carefully crafted fine-grained semantic descriptions, current VLT methods fail to effectively exploit structured semantic guidance, achieving only marginal improvements or even performance degradation (AUC changes of -0.26 to +0.71). This indicates that existing cross-modal fusion strategies fundamentally cannot exploit the rich semantic cues embedded in SOI descriptions. In contrast, we explore a fundamentally different paradigm by employing large-scale vision-language models (VLM) as external cognitive engines for traditional RGB trackers. This approach demonstrates substantial improvements under semantic guidance (AUC gains up to 0.93), significantly outperforming existing VLT methods and providing comprehensive analysis of semantic guidance potential for addressing SOI challenges.

The main contributions of this work are as follows:
(1) We present the first systematic investigation and quantification of SOI's impact on tracking and validate external guidance potential under SOI scenarios.
(2) We construct SOIBench, the first comprehensive semantic cognitive benchmark for SOI challenges.
(3) Using SOIBench, we systematically reveal VLT method limitations and explore VLM-assisted tracking potential, providing methodological guidance for the cognitive evolution of tracking algorithms.


\section{Related Work}

\textbf{SOT Benchmarks.}
Early SOT datasets focused primarily on short-term tracking with limited environmental complexity \cite{10.1109/CVPR.2013.312otb, Huang2021got10k}. Recent datasets have introduced longer sequences and richer scenarios \cite{fan2019lasot, 9720246git}, while visual-language tracking datasets have also emerged \cite{wang2021towards, li2024dtllm}. However, existing benchmarks lack systematic consideration of SOI challenges and do not provide dedicated annotations for fine-grained SOI analysis. This gap has made it difficult to isolate and study the specific impact of similar object interference on tracking performance \cite{soi2023}. To address this limitation, we construct SOIBench, which automatically mines SOI scenarios through multi-tracker consensus and provides hierarchical semantic annotations for SOI challenges.

\textbf{SOT Methods.}
Modern SOT methods are predominantly based on Siamese network architectures \cite{bertinetto2016fullysiamfc, voigtlaender2020siam}, with recent advances incorporating transformer designs \cite{ye2022joint, wei2023autoregressive, zheng2024odtrack, chen2025sutrack}.  Some methods \cite{voigtlaender2020siam, mayer2021learning, kou2023zoomtrack, 2025tcsvtMFDSTrack} have begun to address robustness under similar object interference. However, existing trackers achieve tracking by image-pair matching that heavily relies on appearance features and lacks semantic discriminative capabilities. In this work, we explore how external semantic guidance can complement traditional visual features to enhance discriminative capabilities under SOI scenarios.


\textbf{Vision-Language Tracking.}
Recent VLT methods \cite{zhou2023joint, zheng2023towardmmtrack, ma2024unifyingUVLTrack, li2025dynamicdutrack, 2025ATCTrack} incorporate textual descriptions to enhance tracking performance through feature fusion strategies. However, these approaches typically employ shallow fusion strategies that cannot effectively exploit complex semantic information for addressing SOI scenarios. With the emergence of large-scale vision-language models like CLIP \cite{radford2021learning}, BLIP \cite{li2022blip}, GPT-4V \cite{openai2024gpt4technicalreport}, and Qwen-VL \cite{wang2024qwen2}, new opportunities arise for more sophisticated semantic understanding. Despite their remarkable cross-modal capabilities, the potential of these advanced VLMs as external cognitive engines for tracking has not been systematically explored. Our work investigates how different semantic guidance approaches—from traditional VLT methods to VLM-assisted frameworks—perform under SOI challenges, providing insights into their respective capabilities and limitations.


\section{Preliminaries}
\label{pre-exp}

As the first systematic investigation to quantify SOI's impact, we design a controlled Online Interference Masking (OIM) experiment. The core idea is simple yet powerful: by eliminating interference sources in a controlled manner, we can directly measure the extent to which SOI constrains tracking performance and establish the ideal upper bound achievable through external guidance.


\textbf{Experimental Setup.} 
We integrate OIM into several state-of-the-art trackers and evaluate on LaSOT \cite{fan2019lasot}. When tracking drift is detected (IoU with GT drops below a threshold), we extract high-confidence candidate boxes from the tracker's confidence map and apply grayscale masking to these regions in the next frame while restoring the target region. This simulates idealized external guidance that eliminates SOI in real-time.

\textbf{Findings and Implications.} As shown in Tab.~\ref{tab:soi_mask_results}, all tested trackers achieve substantial performance gains (AUC gains of 2.31-4.35), directly confirming SOI's central role in tracking failures. However, Fig.~\ref{fig:soi_mask_vis}(b) reveals that trackers quickly revert to failure modes once guidance is removed, exposing their fundamental reliance on appearance matching and lack of semantic-level cognitive understanding of target identity. These results establish two key insights: (1) SOI is a primary factor limiting tracking robustness, and (2) external guidance has significant potential to help trackers address SOI challenges. However, mask-based guidance involves GT leakage impractical for deployment, which motivates our exploration of semantic textual guidance as a practical alternative that can provide equivalent cognitive assistance. Detailed experimental results and additional visualizations are provided in the App.A.



\begin{table}[t]
\centering
\setlength{\tabcolsep}{4pt}
\small
\begin{tabular}{lccc}
\toprule
\textbf{Method} & \textbf{Baseline} & \textbf{+OIM} & \textbf{$\Delta$} \\
\midrule
OSTrack-B \cite{ye2022joint} & 70.33 & \textbf{74.68} & +4.35 \\
ODTrack-L \cite{zheng2024odtrack} & 73.86 & \textbf{77.70} & +3.84 \\
LoRAT-L \cite{lorat} & 73.10 & \textbf{75.41} & +2.31 \\
SUTrack-L \cite{chen2025sutrack} & 74.64 & \textbf{78.42} & +3.78 \\
\bottomrule
\end{tabular}
\caption{Performance improvement with Online Interference Masking (OIM) on LaSOT, report AUC.}
\label{tab:soi_mask_results}
\end{table}



\begin{figure*}[t]
    \centering
    \includegraphics[width=\linewidth]{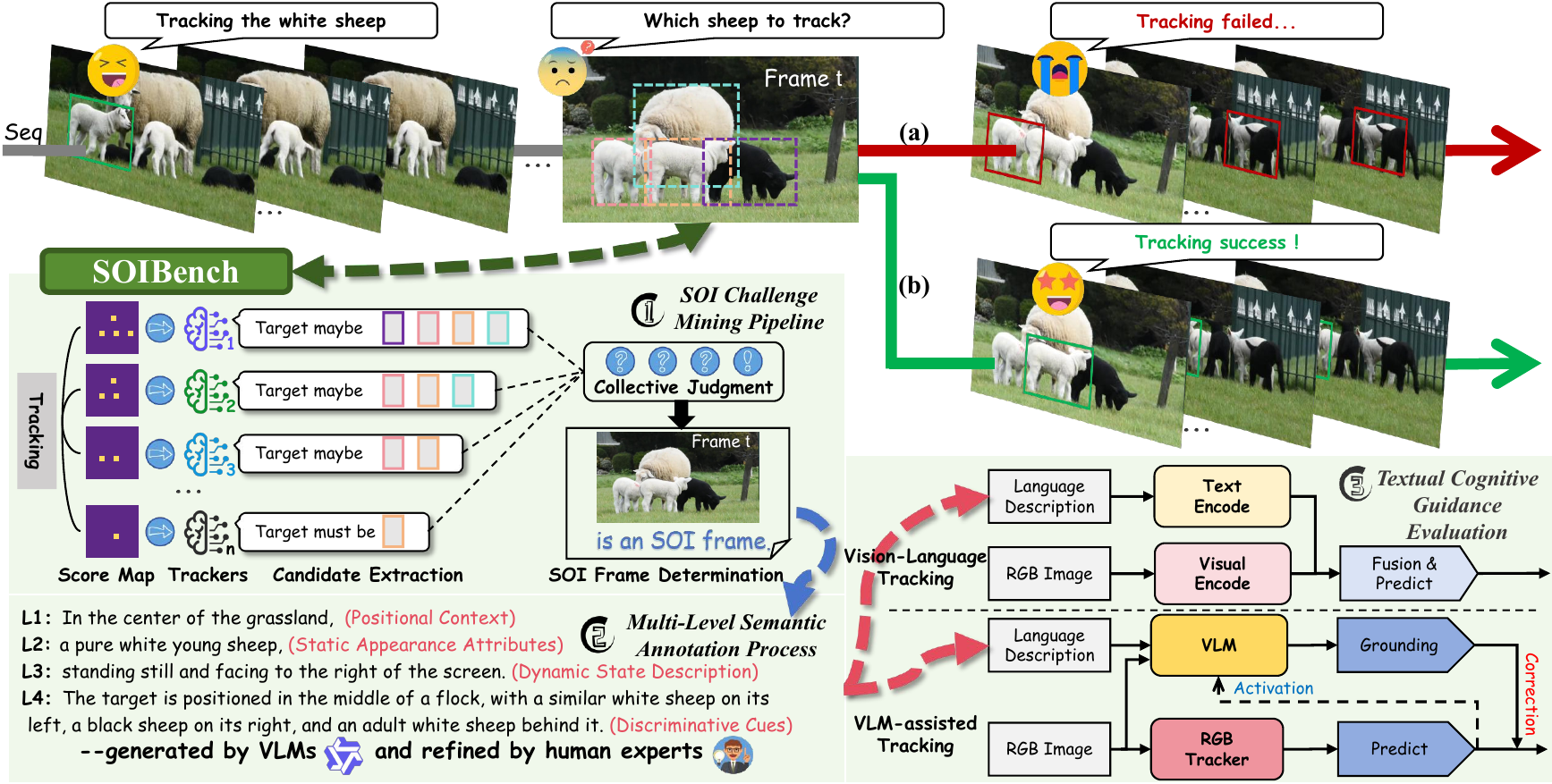}
    \caption{Overview of the SOIBench framework. SOIBench integrates an SOI frame mining pipeline that determines challenging frames through multi-tracker collective judgment, a multi-level semantic annotation process that generates hierarchical textual descriptions for SOI frames, and an SOT evaluation environment with textual cognitive guidance. Arrows (a) and (b) illustrate traditional tracking pipeline and SOIBench-enabled textual cognitive guidance tracking pipeline, respectively.}
    \label{fig:soibench_overview}
\end{figure*}

\section{SOIBench}

In this section, we propose \textbf{SOIBench}---the first comprehensive benchmark specifically designed for SOI challenges with integrated semantic guidance. 
Rather than only providing datasets, SOIBench offers a complete framework covering data mining, semantic annotation, and systematic evaluation. 
As shown in Fig.~\ref{fig:soibench_overview}, it includes three core components: (1) an automated SOI mining pipeline based on multi-tracker collective judgment, (2) a multi-level textual annotation protocol inspired by cognitive linguistics, and (3) a dual-paradigm evaluation environment for systematic assessment of semantic guidance approaches.

\subsection{SOI Challenge Mining Pipeline}
Traditional manual annotation of SOI challenges suffers from subjectivity and high costs. Our automated pipeline instead leverages multi-tracker collective judgment to objectively identify challenging scenarios, ensuring that SOI frames correspond precisely to current trackers' cognitive boundaries. The core insight is that confidence score maps reflect trackers' perceived target position probability distribution within the search region ~\cite{mayer2021learning}. When multiple advanced trackers simultaneously produce high-confidence multi-peak responses on the same frame, it indicates similar object interference causing generalized algorithmic confusion.

Guided by this observation, we design a structured three-phase workflow to reliably mine SOI frames while balancing precision and efficiency. It consists of: (1) extracting high-confidence candidate boxes from each tracker for frame $t$; (2) applying a voting mechanism to determine whether $t$ constitutes an SOI frame based on these candidate sets and tracker status; and (3) performing sequence-level optimization to refine the identified SOI frames.




\textbf{Phase 1. Candidate Extraction.}
For each frame $t$, we extract significant peaks from tracker $i$'s confidence score map $H_t^i \in \mathbb{R}^{W \times H}$ through local maximum suppression and adaptive thresholding:
\begin{equation}
\begin{aligned}
P_t^i(x, y) &= \mathrm{MaxPool}_{k \times k}(H_t^i), \\
\text{Peak}^i(x, y) &= 
\begin{cases}
1, & \text{if }
\begin{array}{l}
P_t^i(x, y) \geq \alpha M_t^i \\
\text{and } P_t^i(x, y) = H_t^i(x, y)
\end{array}
\\[2pt]
0, & \text{otherwise}
\end{cases}
\end{aligned}
\end{equation}

\noindent
where $k \times k$ denotes the pooling kernel size, $\alpha = 0.6$, and $M_t^i = \max(P_t^i)$. This operation effectively filters out secondary responses and highlights primary candidate positions. Valid peaks are decoded into bounding boxes and combined with ground-truth to form candidate sets after IoU-based deduplication:
\begin{equation}
\begin{aligned}
\mathcal{C}_t^{i,\text{raw}} &= \{GT_t\} \cup \mathrm{Track}(H_t^i(\text{Peak}^i(x,y))), \\
\mathcal{C}_t^i &= \{ c \in \mathcal{C}_t^{i,\text{raw}} \mid \forall c' \neq c, \mathrm{IoU}(c, c') \leq \beta \}
\end{aligned}
\end{equation}
\noindent
where $\beta = 0.4$ is the IoU threshold for deduplication. $\mathcal{C}_t^i$ is the final candidate set, containing the ground-truth as the first element, followed by any detected distractor candidates.

\textbf{Phase 2. SOI Frame Determination.}
Based on the candidate sets $\mathcal{C}_t^i$ from each tracker, we employ a multi-tracker voting mechanism to identify SOI frames through collective consensus. Each tracker's voting eligibility depends on its tracking status, which reflects both prediction accuracy and the presence of multiple candidates:
(1) \textbf{Correct}: $\mathrm{IoU}(\text{pred}_t^i, GT_t) \geq \tau$ and $|\mathcal{C}_t^i| = 1$ (accurate tracking, no confusion);
(2) \textbf{Compromise}: $\mathrm{IoU}(\text{pred}_t^i, GT_t) \geq \tau$ and $|\mathcal{C}_t^i| > 1$ (correct but confused);
(3) \textbf{Drift}: $\mathrm{IoU}(\text{pred}_t^i, GT_t) < \tau$ and $|\mathcal{C}_t^i| > 1$ (failed and confused);
(4) \textbf{Fail}: $\mathrm{IoU}(\text{pred}_t^i, GT_t) < \tau$ and $|\mathcal{C}_t^i| = 1$ (failed but confident);
where $\tau = 0.6$. Trackers with Correct status cast negative votes $(V_{t,i} = 0)$, while trackers with multiple candidates (Compromise or Drift) cast positive votes $(V_{t,i} = 1)$. Then $t$ is marked as SOI when the majority vote positively:
\begin{equation}
\text{SOI}_t = \mathbb{1}\left(\textstyle\sum_{i=1}^{N} V_{t,i} \geq \left(\lfloor N/2 \rfloor + 1\right)\right)
\end{equation}

\noindent
where $N$ is the total number of trackers.

\textbf{Phase 3. Sequence-Level Optimization.}
After obtaining all frame-level SOI determinations, we perform sequence-level optimization to ensure annotation efficiency and quality. This process addresses two key aspects: temporal redundancy and annotation suitability.

To avoid redundant annotations caused by temporal persistence of SOI challenges, we implement temporal filtering with 30-frame intervals.
After identifying an SOI frame at time $t$, we skip the next 29 frames unless significant scene changes occur, determined by substantial target state changes.
Additionally, we filter out frames with targets too small for accurate annotation. The final SOI frame set $\mathcal{S}$ combines these filtering criteria:
\begin{equation}
\begin{split}
\mathcal{S} = \{ t : &\text{SOI}_t = 1 \land  \frac{\mathrm{Area}(GT_t)}{\mathrm{Area}(I_t)} \geq 0.001 \land \\
(t - t_{last} \geq 30& \lor \mathrm{IoU}(GT_{t}, GT_{t_{last}}) < \sigma_1)  \},
\end{split}
\end{equation}
\noindent
where $\sigma_1 = 0.5$, $t_{last}$ represents the most recently selected SOI frame. This ensures selected SOI frames are both representative of genuine challenges and suitable for high-quality semantic annotation.

\subsection{Multi-Level Semantic Annotation Process}

To provide precise semantic guidance for selected SOI frames, we design a hierarchical annotation protocol grounded in cognitive linguistics principles. Unlike conventional tracking annotations that provide only basic object descriptions, our protocol follows two key design principles: \emph{description concreteness} for accurate target specification and \emph{discriminative saliency} for distinguishing targets from similar distractors.

Our four-level annotation hierarchy progressively refines semantic understanding:
(1) \textbf{L1---Positional Context}: Spatial relationships and environmental location;
(2) \textbf{L2---Static Appearance Attributes}: Visual characteristics and distinctive attributes; 
(3) \textbf{L3---Dynamic State Description}: Motion patterns and behavioral descriptions; 
(4) \textbf{L4---Discriminative Cues}: Contextual differences from similar objects.
This progressive refinement from basic spatial information to fine-grained discriminative features enables comprehensive target-focused understanding tailored to SOI scenarios. To balance annotation quality and efficiency, we employ a hybrid approach: VLMs generate initial descriptions across all four levels, which are subsequently refined and validated by trained human annotators through systematic quality control.
Further details can be found in App.B.

\subsection{Evaluation Environment}
\label{4.3}
SOIBench establishes a standardized evaluation environment that enables systematic assessment of semantic guidance approaches under controlled SOI scenarios. As illustrated in Fig.~\ref{fig:soibench_overview}, our evaluation accommodates two complementary paradigms that represent different philosophies for integrating semantic information into tracking.

\textbf{Vision-Language Tracking (VLT) Evaluation.} This paradigm assesses existing VLT methods that integrate textual descriptions through multimodal fusion strategies. These methods process visual and textual inputs simultaneously during inference, embedding semantic information directly into the tracking pipeline for enhanced target discrimination.

\textbf{VLM-Assisted Tracking Evaluation.} We explore a novel tracking paradigm where large-scale Vision-Language Models (VLM) serve as external cognitive engines for traditional RGB trackers. VLMs are activated conditionally---only when tracker confidence drops below a threshold and SOI annotations are available---performing visual grounding tasks to correct erroneous predictions.

This dual-paradigm evaluation enables a comprehensive comparison of different semantic integration strategies, revealing their respective capabilities and limitations when confronting SOI challenges. By evaluating both embedded and external semantic guidance approaches, SOIBench provides insights into optimal architectures for semantic-aware tracking systems.

\begin{table*}[!t]
\centering
\small
\setlength{\tabcolsep}{3pt}
\begin{tabular}{lcccccccccccc}
\toprule
\multirow{2}{*}{\textbf{Model}} & \multirow{2}{*}{\makecell{Visual\\Processing}} & \multirow{2}{*}{\makecell{Text\\Processing}} & \multirow{2}{*}{\makecell{Image\\Size}} &\multicolumn{3}{c}{\textbf{AUC}} & \multicolumn{3}{c}{\textbf{$\text{P}_{\text{N}}$}} & \multicolumn{3}{c}{\textbf{P}} \\
 \cmidrule(lr){5-7} \cmidrule(lr){8-10} \cmidrule(lr){11-13}
&   &  &  & Base & SOI$_{30}$ & SOI$_1$ & Base & SOI$_{30}$ & SOI$_1$ & Base & SOI$_{30}$ & SOI$_1$ \\
\midrule
\multicolumn{13}{l}{\textbf{\textit{Vision-Language Trackers}}} \\
JointNLT \cite{zhou2023joint} & Swin-B & BERT & 320 & 56.87 & \cellcolor{lightgreen}57.09 & \cellcolor{lightgreen}57.08 & 65.87 & \cellcolor{lightcoral}65.76 & \cellcolor{lightgreen}66.10 & 59.21 & \cellcolor{lightcoral}59.02 & \cellcolor{lightgreen}59.41 \\
All-in-One \cite{zhang2023allallinone} & ViT-B & Tokenizer & 256 & 72.18 & \cellcolor{lightgreen}72.51 & \cellcolor{lightgreen}72.47 & 82.53 & \cellcolor{lightgreen}82.90 & \cellcolor{lightgreen}82.86 & 79.69 & \cellcolor{lightgreen}80.04 & \cellcolor{lightgreen}80.03 \\
MMTrack \cite{zheng2023towardmmtrack}& ViT-B & RoBERTa-B & 384 & 69.87 & \cellcolor{lightcoral}69.73 & \cellcolor{lightcoral}69.78 & 82.14 & \cellcolor{lightcoral}81.92 & \cellcolor{lightcoral}81.97 & 75.61 & \cellcolor{lightcoral}75.39 & \cellcolor{lightcoral}75.42 \\
UVLTrack-B \cite{ma2024unifyingUVLTrack}& ViT-B & BERT &  256 & 68.12 & \cellcolor{lightgreen}68.14 & \cellcolor{lightcoral}67.91 & 79.43 & \cellcolor{lightgreen}79.45 & \cellcolor{lightcoral}79.24 & 73.42 & \cellcolor{lightgreen}73.56 & \cellcolor{lightcoral}73.30 \\
UVLTrack-L \cite{ma2024unifyingUVLTrack}& ViT-L & BERT & 256 & 70.63 & \cellcolor{lightgreen}71.34 & \cellcolor{lightgreen}70.96 & 82.33 & \cellcolor{lightgreen}83.08 & \cellcolor{lightgreen}82.68 & 77.77 & \cellcolor{lightgreen}78.52 & \cellcolor{lightgreen}78.14 \\
DuTrack-B \cite{li2025dynamicdutrack}& ViT-B & Tokenizer & 224 & 72.83 & \cellcolor{lightgreen}72.87 & \cellcolor{lightgreen}72.96 & 83.65 & \cellcolor{lightgreen}83.69 & \cellcolor{lightgreen}83.80 & 80.92 & \cellcolor{lightgreen}80.96 & \cellcolor{lightgreen}81.06 \\
SUTrack-B \cite{chen2025sutrack}& ViT-B & CLIP-L & 384 & 71.10 & \cellcolor{lightgreen}71.38 & \cellcolor{lightgreen}71.33 & 82.90 & \cellcolor{lightgreen}83.20 & \cellcolor{lightgreen}83.11 & 79.03 & \cellcolor{lightgreen}79.24 & \cellcolor{lightgreen}79.24 \\
SUTrack-L \cite{chen2025sutrack}& ViT-L & CLIP-L & 384 & 70.83 & \cellcolor{lightgreen}71.34 & \cellcolor{lightgreen}71.41 & 81.88 & \cellcolor{lightgreen}82.45 & \cellcolor{lightgreen}82.42 & 78.10 & \cellcolor{lightgreen}78.66 & \cellcolor{lightgreen}78.67 \\
ATCTrack-B \cite{2025ATCTrack}& ViT-B & BERT & 384 & 73.88 & \cellcolor{lightcoral}73.62 & \cellcolor{lightcoral}73.77 & 86.73 & \cellcolor{lightcoral}86.26 & \cellcolor{lightcoral}86.55 & 81.56 & \cellcolor{lightcoral}81.28 & \cellcolor{lightcoral}81.41 \\
ATCTrack-L \cite{2025ATCTrack}& ViT-L & BERT & 384 & 74.53 & \cellcolor{lightgreen}74.83 & \cellcolor{lightgreen}74.64 & 86.99 & \cellcolor{lightgreen}87.33 & \cellcolor{lightgreen}87.13 & 82.14 & \cellcolor{lightgreen}82.43 & \cellcolor{lightcoral}82.13 \\
\midrule
\multicolumn{2}{l}{\textbf{\textit{VLM-Assisted Trackers}}} \\
OSTrack \cite{ye2022joint}& ViT-B & Qwen2.5-VL† & 384 & 70.33 & \cellcolor{lightgreen}71.30 & \cellcolor{lightgreen}71.26 & 79.98 & \cellcolor{lightgreen}81.29 & \cellcolor{lightgreen}81.33 & 76.62 & \cellcolor{lightgreen}77.71 & \cellcolor{lightgreen}77.79 \\
SeqTrack-B \cite{chen2023seqtrack}& ViT-B & Qwen2.5-VL† & 384 & 71.53 & \cellcolor{lightgreen}71.95 & \cellcolor{lightgreen}71.64 & 81.05 & \cellcolor{lightgreen}81.70 & \cellcolor{lightgreen}81.26 & 77.83 & \cellcolor{lightgreen}78.21 & \cellcolor{lightgreen}77.85 \\
SeqTrack-L \cite{chen2023seqtrack}& ViT-L & Qwen2.5-VL† & 384 & 72.51 & \cellcolor{lightgreen}73.16 & \cellcolor{lightgreen}72.75 & 81.53 & \cellcolor{lightgreen}82.28 & \cellcolor{lightgreen}81.84 & 79.25 & \cellcolor{lightgreen}80.05 & \cellcolor{lightgreen}79.55 \\
ODTrack-B \cite{zheng2024odtrack} & ViT-B & Qwen2.5-VL† & 384 & 72.85 & \cellcolor{lightcoral}72.83	& \cellcolor{lightgreen}73.01 & 82.76 & \cellcolor{lightgreen}82.87 & \cellcolor{lightgreen}82.97 & 80.16 & \cellcolor{lightcoral}79.97 & \cellcolor{lightgreen}80.35 \\
ODTrack-L \cite{zheng2024odtrack} & ViT-L & Qwen2.5-VL† & 384 & 73.86 & \cellcolor{lightgreen}74.33 & \cellcolor{lightgreen}74.19 & 84.07 & \cellcolor{lightgreen}84.65 & \cellcolor{lightgreen}84.47 & 82.26 & \cellcolor{lightgreen}82.80 & \cellcolor{lightgreen}82.66 \\
SUTrack-B \cite{chen2025sutrack} & ViT-B & Qwen2.5-VL† & 384 & 73.81 & \cellcolor{lightgreen}73.98 & \cellcolor{lightgreen}73.85 & 83.29 & \cellcolor{lightgreen}84.14	 & \cellcolor{lightgreen}83.47 & 81.39 & \cellcolor{lightgreen}81.41	 & \cellcolor{lightgreen}81.57 \\
SUTrack-L \cite{chen2025sutrack} & ViT-L & Qwen2.5-VL† & 384 & 74.64 & \cellcolor{lightgreen}74.97 & \cellcolor{lightgreen}74.66 & 84.13 & \cellcolor{lightgreen}84.41 & \cellcolor{lightgreen}84.15 & 82.43 & \cellcolor{lightgreen}82.78 & \cellcolor{lightgreen}82.56 \\
\bottomrule
\end{tabular}
\caption{Comprehensive evaluation on $\text{LaSOT}_{\text{SOI}}$. \colorbox{lightgreen}{Green} indicates improvement, \colorbox{lightcoral}{red} indicates degradation. † indicates external VLM assistance for SOI guidance.}
\label{tab:comprehensive_evaluation}
\end{table*}

\section{Experiments and Analysis}



This section evaluates the effectiveness of semantic guidance in addressing SOI challenges. We pursue three objectives: (1) demonstrate SOIBench's applicability by constructing LaSOT$_{\text{SOI}}$, (2) evaluate two tracking paradigms under semantic guidance—existing VLT methods and our proposed VLM-assisted approaches, and (3) perform diagnostic analysis to reveal whether current methods truly exploit SOI semantic guidance, exposing fundamental limitations and providing insights for future development.

\subsection{Experimental Details}

\begin{figure}[!t]
\centering
\includegraphics[width=\linewidth]{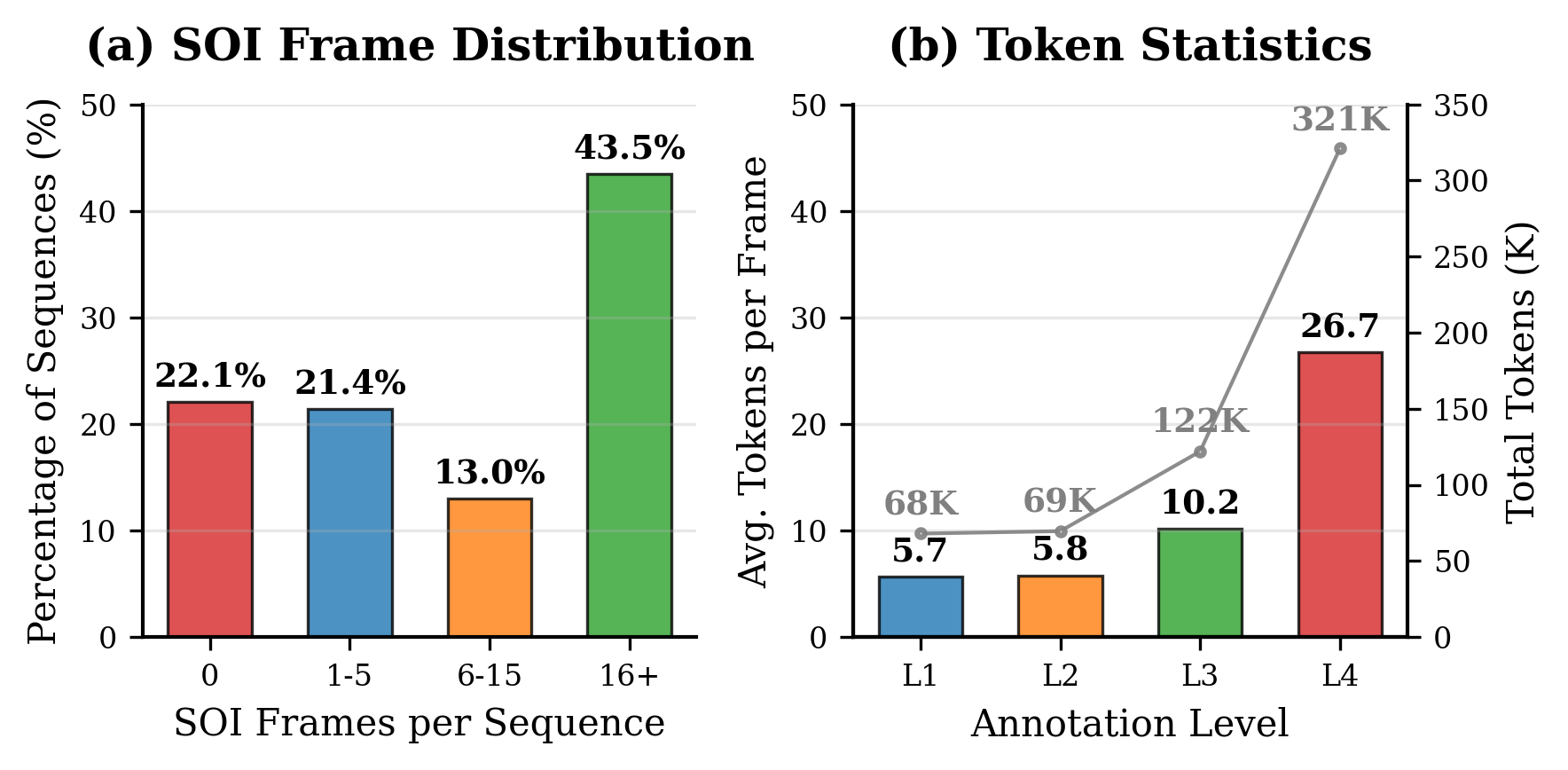}
\caption{SOIBench construction statistics. (a) SOI challenge mining statistics; (b) Semantic annotation statistics.}
\label{fig:soi_stats}
\end{figure}

\textbf{Dataset.}
To validate SOIBench's effectiveness, we instantiate it on LaSOT~\cite{fan2019lasot}, a representative long-term tracking dataset known for its complex scenarios and extended sequences. We employ four SOTA trackers \cite{ye2022joint, zheng2024odtrack, lorat, chen2025sutrack} as consensus judges to automatically mine SOI frames and generate hierarchical semantic annotations following our proposed protocol.

\textit{Mining Results.} Our pipeline successfully extracts 12K+ high-quality SOI challenge frames from the original 685K frames, achieving a selective ratio of ~2\% that prioritizes quality over quantity. Notably, 43.5\% of sequences contain more than 15 SOI frames, confirming the pervasive nature of SOI challenges in real-world long-term tracking scenarios  (Fig.~\ref{fig:soi_stats}a). 

\textit{Semantic Annotation Statistics.} The annotation process generates approximately 580k tokens across all semantic levels (Fig.~\ref{fig:soi_stats}b). Significantly, L4 (discriminative contextual cues) contributes the largest portion with 321k tokens, reflecting the complexity required to distinguish targets from similar distractors---a finding that underscores the cognitive demands of SOI scenarios.

The resulting $\text{LaSOT}_{\text{soi}}$ dataset demonstrates SOIBench's generalizability and adaptability. SOIBench can be applied to any SOT dataset for SOI benchmark construction and can evolve dynamically with algorithmic advances by automatically filtering out frames that become trivial for emerging SOTA methods, ensuring continued evaluation relevance.

\textbf{Evaluation Methodology.}
We evaluate two tracking paradigms under the SOIBench framework, corresponding to the designs introduced in Sec.~\ref{4.3}. 
Specifically, we assess: 
\textbf{(1) VLT Methods:} which integrates visual and textual inputs through multimodal fusion; and \textbf{(2) VLM-Assisted RGB Trackers:} where an excellent VLM Qwen2.5-VL-32B~\cite{wang2024qwen2} is conditionally engaged as an external cognitive module to assist RGB trackers. The specific configurations of selected representative methods are detailed in Tab.~\ref{tab:comprehensive_evaluation}.

\textbf{Implementation Details.}
For fair evaluation and analysis, all methods use officially released weights without fine-tuning. 
We evaluate three settings: \textbf{Base} represents VLT methods using LaSOT's official text and VLM-assisted methods performing pure RGB tracking; \textbf{SOI$_{30}$} and \textbf{SOI$_1$} introduce our SOI descriptions with each SOI text maintaining a validity period of 30 and 1 frame, respectively. VLT models fall back to official text when SOI descriptions expire, while VLM-assisted methods use the VLM module only within the validity window, and revert to RGB tracking thereafter.
This experimental design allows us to analyze both the immediate and sustained effects of semantic guidance.
All experiments are conducted on a server with 8 RTX 3090 GPUs, reporting metrics: Area Under Curve (AUC), Precision (P), and Normalized Precision ($P_{\text{Norm}}$).


\subsection{Performance Evaluation under SOI Guidance}

Following the dual-paradigm evaluation framework established in Sec.~\ref{4.3}, Tab.~\ref{tab:comprehensive_evaluation} reveals notable differences in how Vision-Language Tracking and VLM-Assisted approaches respond to semantic guidance on LaSOT$_{\text{SOI}}$.




\textbf{Vision-Language Tracking Results.}
Despite carefully crafted textual descriptions, most VLT methods achieve only marginal gains under SOI guidance (0.2–0.7 AUC points), while certain models (e.g., MMTrack, ATCTrack-B) even show slight performance degradation.
Notably, VLT algorithms perform better under the SOI$_{1}$ setting than SOI$_{30}$, which we attribute to the fact that extending the validity of a single description across 30 frames does not provide sustained guidance but instead disrupts feature modeling, thereby reducing effectiveness.

We attribute this limited or even negative impact to two inherent deficiencies of the VLT paradigm. First, textual features are projected into low-dimensional latent spaces before fusion, which inevitably discards a substantial amount of semantic cognitive cues. Second, VLT models are trained primarily on relatively small-scale SOT datasets with overly simplistic textual annotations, leaving them severely under-equipped with the semantic reasoning ability required to exploit SOI guidance effectively. 
These results demonstrate that despite attempts to integrate language modalities, current VLT methods remain fundamentally constrained by their reliance on visual features and cannot overcome the intrinsic limitations of appearance-based tracking.



\textbf{VLM-Assisted Tracking Results.}
In contrast, our proposed VLM-assisted paradigm demonstrates more stable and substantial improvements. On the representative RGB trackers we evaluate, assistance from VLM consistently yields performance gains, with AUC improvements ranging from 0.4 to 1.0. Notably, this paradigm exhibits the opposite temporal behavior: it shows greater benefits under the SOI$_{30}$ setting compared to SOI$_{1}$, as the longer validity period provides more opportunities to activate the VLM module and thus offers greater potential for correction.


The consistent improvements across different tracker architectures validate our core hypothesis: external semantic cognition can effectively address SOI challenges when properly integrated. The VLM's pretraining on large-scale vision-language corpora enables sophisticated semantic reasoning to interpret SOI guidance, while the cooperative design preserves RGB tracking efficiency by engaging external cognition only at critical moments. These results not only highlight the promising potential of VLM-assisted tracking but also establish a new paradigm for integrating large-scale vision-language models into SOT tasks.


\begin{table}[htbp]
\centering
\small
\setlength{\tabcolsep}{6pt}
\begin{tabular}{lccc}
\toprule
\textbf{Method} & \textbf{Base} & \textbf{$\text{SOI}_{30}$} & \textbf{Noise} \\
\midrule
JointNLT \cite{zhou2023joint} & 56.87 & 57.09 & 56.42 \\
All-in-One \cite{zhang2023allallinone} & 72.18 & 72.51 & 72.11 \\
MMTrack \cite{zheng2023towardmmtrack} & 69.87 & 69.73 & 69.71 \\
UVLTrack-L \cite{ma2024unifyingUVLTrack} & 70.63 & 71.34 & 70.56 \\
DuTrack-B \cite{li2025dynamicdutrack} & 72.83 & 72.87 & 72.81 \\
SUTrack-L \cite{chen2025sutrack} & 70.83 & 71.34 & 71.41 \\
ATCTrack-L \cite{2025ATCTrack} & 74.53 & 74.83 & 74.92 \\
\bottomrule
\end{tabular}
\caption{Comparative evaluation of VLT methods with different semantic text inputs, reporting AUC.}
\label{tab:noise_control}
\end{table}

\subsection{Beyond Performance: Diagnosing SOI Comprehension}
To further assess how well the two tracking paradigms understand and leverage SOI textual guidance, we design two independent diagnostic experiments. For VLT methods, we employ semantic perturbation to examine their sensitivity to textual content; for the VLM-assisted paradigm, we conduct a separate grounding evaluation to test the effectiveness of the introduced VLM.



\textbf{Semantic Perturbation: Do VLT Methods Understand Text?}
We construct controlled perturbation experiments by replacing original SOI descriptions with semantically mismatched ``noise" text that retains approximately 50\% of original tokens while conveying contradictory meanings. This design isolates semantic comprehension from superficial token-level dependencies.

Tab.~\ref{tab:noise_control} reveals a striking finding: most VLT models show negligible performance degradation under noisy inputs, with some (SUTrack-L, ATCTrack-L) even improving slightly. This counterintuitive result suggests that current VLT architectures rely predominantly on visual processing, treating textual input more as passive auxiliary information rather than active discriminative guidance. The apparent insensitivity to semantic corruption exposes fundamental limitations in cross-modal understanding capabilities.


\begin{table}[!t]
\centering
\setlength{\tabcolsep}{2pt}
\small
\begin{tabular}{lccc}
\toprule
\textbf{Test Object} & \textbf{SR@50} & \textbf{SR@75} & \textbf{Mean IoU} \\
\midrule
\textbf{Human} & \textbf{0.725} & \textbf{0.455} & \textbf{0.664} \\
Qwen2.5VL-32B \cite{wang2024qwen2} & 0.479 & 0.223 & 0.497 \\
Qwen2.5VL-7B \cite{wang2024qwen2} & 0.476 & 0.265 & 0.471 \\
Qwen2.5VL-3B \cite{wang2024qwen2} & 0.397 & 0.249 & 0.420 \\
DeepSeek-VL2 \cite{wu2024deepseekvl2} & 0.110 & 0.056 & 0.140 \\
\bottomrule
\end{tabular}
\caption{Grounding performance of VLMs and humans using SOI descriptions.}
\label{tab:vlm_grounding_auc}
\end{table}

\textbf{Grounding Evaluation: Can VLMs Follow SOI Instructions?}
We conduct direct visual grounding experiments on SOI frames using our semantic guidance descriptions. This test isolates semantic understanding from tracking-specific optimizations by evaluating VLMs' ability to localize targets based solely on textual descriptions.

Tab.~\ref{tab:vlm_grounding_auc} shows that humans achieve optimal performance (Mean IoU 0.664), validating the quality of our textual descriptions. Among VLMs, Qwen2.5VL-32B performs best (Mean IoU 0.497) but remains notably below human-level understanding. While VLMs demonstrate meaningful grounding capabilities, they struggle with complex scenarios involving multiple distractors or ambiguous contexts. This explains why practical semantic guidance cannot achieve the substantial gains observed with mask-based guidance, and further reveals the gap between current VLM capabilities and the cognitive reasoning required for robust SOI resolution. Please see App.D for more details.


\begin{figure}[!t]
    \centering
    \includegraphics[width=\linewidth]{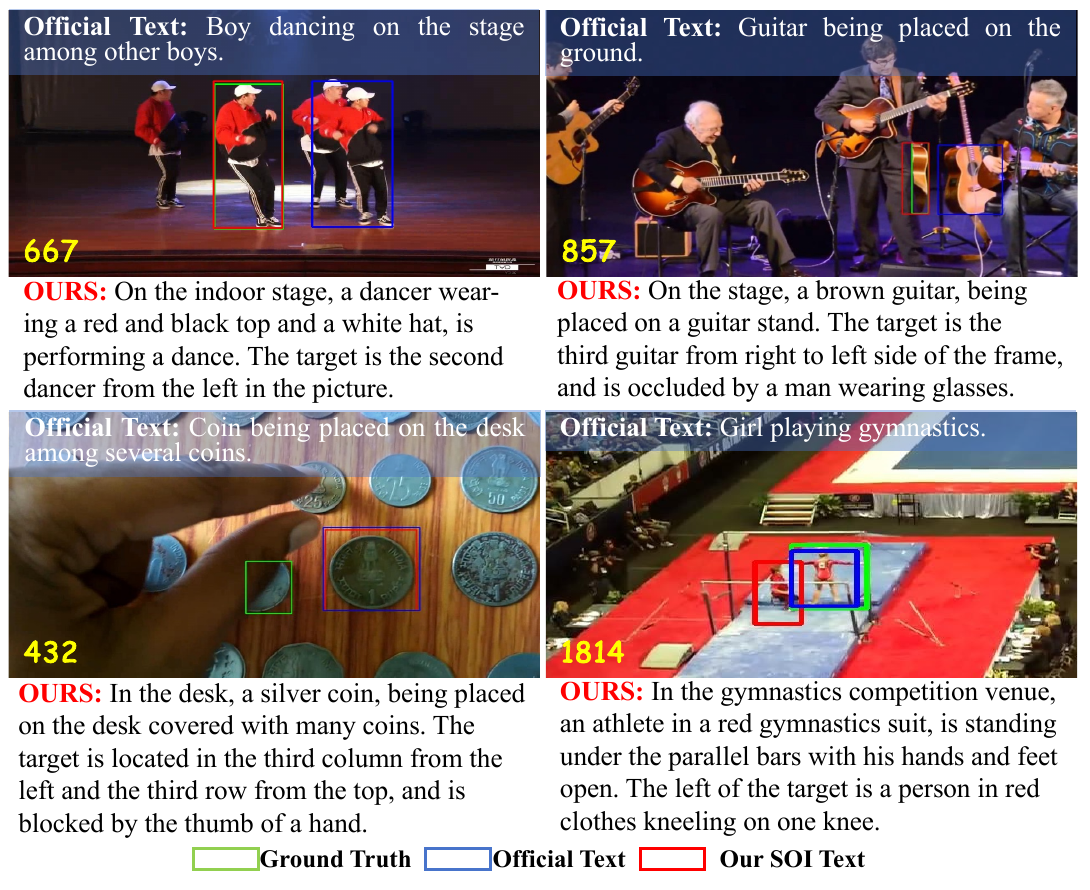}
    \caption{Qualitative comparison results of MMTrack with Official text and our SOI text.}
    \label{fig:vlt_visualization}
\end{figure}

\begin{figure}[!t]
    \centering
    \includegraphics[width=\linewidth]{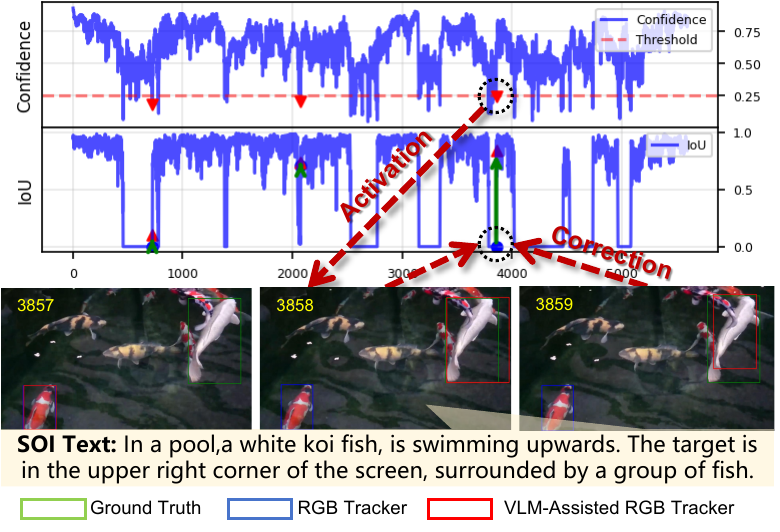}
    \caption{Visualization of OSTack assisted by VLM.}
    \label{fig:vlm_visualization}
\end{figure}

\subsection{Qualitative Analysis}

We provide qualitative visualization of both tracking paradigms under SOI challenges.

Fig.~\ref{fig:vlt_visualization} demonstrates representative cases of UVLTrack. The upper examples show successful cases where VLT can effectively leverage our hierarchical SOI descriptions to maintain accurate tracking despite visual similarity between targets and distractors. However, the lower examples reveal significant limitations: even when provided with detailed semantic descriptions that clearly distinguish targets from surrounding similar objects, VLT methods fail to utilize this guidance effectively. This highlights that current VLT architectures struggle to establish meaningful correspondence between fine-grained textual descriptions and visual scene understanding, confirming our findings that these methods lack sophisticated semantic comprehension capabilities.

Fig.~\ref{fig:vlm_visualization} illustrates the corrective capability of VLM-assisted tracking. When SOI text is available and the tracker's confidence falls below the threshold, the VLM module is activated and successfully corrects tracking errors. As can be further observed, the VLM-assisted tracker (red box) maintains stable and accurate tracking in subsequent frames, while the baseline (blue box) continues to drift. However, this paradigm simultaneously exposes a critical limitation: when the tracker generates high-confidence predictions on incorrect targets, the VLM module cannot be activated, resulting in missed opportunities for crucial corrections. It presents an urgent evolutionary need—to construct a semantic verification mechanism independent of confidence assessment, enabling more reliable SOI tracking.

\section{Conclusion}
This paper presents the first systematic investigation of Similar Object Interference (SOI) as a critical bottleneck in visual tracking. Through controlled experiments, we quantitatively demonstrate SOI's central role in tracking failures and validate the potential of external cognitive guidance. We introduce \textbf{SOIBench}, the first comprehensive benchmark for SOI challenges with integrated semantic guidance, encompassing automated mining, multi-level annotation, and standardized evaluation protocols.
Our systematic evaluation reveals striking limitations in existing vision-language tracking methods' semantic comprehension capabilities, while our proposed VLM-assisted paradigm demonstrates consistent improvements by leveraging external cognitive engines. These findings not only establish SOI as a fundamental research direction but also provide essential tools and insights for advancing semantic cognitive tracking research, ultimately contributing to the cognitive evolution of visual tracking algorithms.

\section{Appendix}
\appendix

In this appendix, we provide comprehensive supplementary materials to support and extend the findings presented in the main manuscript:

\begin{itemize}
  \item \textbf{Online Interference Masking (OIM) Controlled Experiment}: Detailed implementation, algorithmic procedures, and comprehensive analysis of the masking mechanism used to quantify SOI impact.
  \item \textbf{SOIBench Framework}: Complete technical specifications of our benchmark construction, including the automated mining pipeline and multi-level semantic annotation protocol.
  \item \textbf{LaSOT$_{\text{SOI}}$ Dataset Construction}: Comprehensive details on dataset instantiation and annotation quality assurance procedures.
  \item \textbf{VLM Grounding Evaluation}: Experimental setup and implementation details for both vision-language models and human annotators.
\end{itemize}

\section{More Details of Online Interference Masking controlled experiment} 
This section provides the implementation details of the Online Interference Masking (OIM) controlled experiment that are omitted from the main manuscript due to space constraints.

\subsection{Online Interference Masking Pipeline}
The OIM mechanism simulates an idealized form of external cognitive guidance to eliminate visual distractions through real-time mask operation, thereby revealing the potential upper limit of the cognitive capabilities of existing trackers. 

OIM adopts a frame-by-frame online processing mode, with the detailed procedure illustrated in Alg. \ref{alg:oim}. The candidate box extraction algorithm maintains consistency with the candidate object extraction mechanism in the main text's SOI challenge mining pipeline, which identifies all local peaks in the confidence map through max-pooling operations and decodes them into corresponding bounding box sets. We design a simple yet effective mask generation strategy: when tracking drift is detected in the current frame, grayscale masking operations are applied to all high-confidence candidate box regions, while simultaneously restoring the ground-truth target region to ensure its visibility, thereby implementing a strong cognitive guidance mechanism.

\begin{algorithm}[htbp]
\caption{Online Interference Masking (OIM) Pipeline}
\label{alg:oim}
\begin{algorithmic}[1]
\REQUIRE Tracker $\mathcal{T}$, Sequence $\{I_1, ..., I_T\}$, Ground-truth boxes $\{GT_1, ..., GT_T\}$, IoU threshold $\tau$
\ENSURE OIM-enhanced predictions $\{\hat{B}_1, ..., \hat{B}_T\}$
\STATE $\hat{B}_1 \gets GT_1$ \hfill \# Initialization in the first frame
\FOR{$t = 2$ to $T$}
    \STATE $I_t^{input} \gets I_t^{''} \text{ if } I_t^{''} \text{ exists, else } I_t$ \hfill \# Enable OIM if available
    \STATE $H_t \gets \mathcal{T}(I_t^{input})$ \hfill \# Run tracker to get confidence score map
    \STATE $\hat{B}_t \gets \textsc{DecodeBox}(H_t)$ \hfill \# Decode predicted box
    \IF{IoU$(\hat{B}_t, GT_t) < \tau$} 
        \STATE \# Drift detected, trigger OIM
        \STATE $\mathcal{C}_t \gets \textsc{Extract}(H_t)$ \hfill \# Extract high-confidence candidate boxes
        \STATE $I_{t+1}^{'} \gets \textsc{Mask}(I_{t+1}, \mathcal{C}_t)$ \hfill \# Mask all candidates
        \STATE $I_{t+1}^{''} \gets \textsc{Fix}(I_{t+1}^{'}, {GT}_{t+1})$ \hfill \# Fix the region of GT
    \ENDIF
\ENDFOR
\RETURN $\hat{B}_1, ..., \hat{B}_T$
\end{algorithmic}
\end{algorithm}

\subsection{Experiment}

Tab. \ref{tab:appendix-soi_mask_results} presents comprehensive comparison results of multiple algorithms on the LaSOT dataset under the OIM mechanism. The results demonstrate consistent and substantial performance improvements across all tested state-of-the-art trackers, providing strong empirical evidence that SOI constitutes a fundamental bottleneck in visual tracking.
All evaluated trackers achieve significant improvements when equipped with OIM guidance. Notably, no tracker experiences performance degradation, confirming the universality of SOI's impact across different architectural designs. The magnitude of improvement varies across tracker families: OSTrack shows the largest improvement (+4.35 AUC), while LoRAT demonstrates smaller but consistent gains (+1.15 AUC average), suggesting different levels of inherent robustness to interference. These results reveal the theoretical upper bounds achievable by current tracking paradigms when SOI constraints are eliminated, with SUTrack-L384 reaching 78.42 AUC under OIM guidance.

\begin{table}[h]
\centering
\small
\renewcommand{\arraystretch}{1.2}
\begin{tabular}{lccc}
\toprule
\textbf{Method} & \textbf{AUC} & \textbf{$P_{\text{Norm}}$} & \textbf{P} \\
\midrule
OSTrack-B \cite{ye2022joint} & 70.33 & 79.98 & 76.62 \\
\hspace{2mm}+OIM & \textbf{74.68} & \textbf{85.64} & \textbf{81.88} \\
\hdashline
\textit{Improvement} & +4.35 & +5.66 & +5.26 \\
\midrule
ODTrack-B \cite{zheng2024odtrack} & 72.85 & 82.76 & 80.16 \\
\hspace{2mm}+OIM & \textbf{75.39} & \textbf{86.06} & \textbf{82.99} \\
ODTrack-L \cite{zheng2024odtrack} & 73.86 & 84.07 & 82.26 \\
\hspace{2mm}+OIM & \textbf{77.70} & \textbf{88.88} & \textbf{86.84} \\
\hdashline
\textit{Avg. Improvement} & +3.19 & +4.06 & +3.71 \\
\midrule
LoRAT-B \cite{lorat} & 71.53 & 81.69 & 79.41 \\
\hspace{2mm}+OIM & \textbf{72.39} & \textbf{82.72} & \textbf{80.43} \\
LoRAT-L \cite{lorat} & 73.10 & 83.80 & 82.63 \\
\hspace{2mm}+OIM & \textbf{75.41} & \textbf{86.00} & \textbf{84.65} \\
LoRAT-G \cite{lorat} & 73.93 & 84.02 & 83.08 \\
\hspace{2mm}+OIM & \textbf{74.62} & \textbf{84.97} & \textbf{83.68} \\
\hdashline
\textit{Avg. Improvement} & +1.15 & +1.37 & +1.22 \\
\midrule
SUTrack-B \cite{chen2025sutrack} & 73.81 & 83.29 & 81.39 \\
\hspace{2mm}+OIM & \textbf{77.90} & \textbf{88.30} & \textbf{86.08} \\
SUTrack-L \cite{chen2025sutrack} & 74.64 & 84.13 & 82.43 \\
\hspace{2mm}+OIM & \textbf{78.42} & \textbf{88.79} & \textbf{87.12} \\
\hdashline
\textit{Avg. Improvement} & +3.94 & +4.84 & +4.69  \\
\bottomrule
\end{tabular}
\caption{Performance improvement with Online Interference Masking (OIM) on LaSOT.}
\label{tab:appendix-soi_mask_results}
\end{table}

\subsection{Qualitative Analysis}

\subsubsection{Successful Correction and Immediate Reversion}

\begin{figure*}[htbp]
    \centering
    \includegraphics[width=1\linewidth]{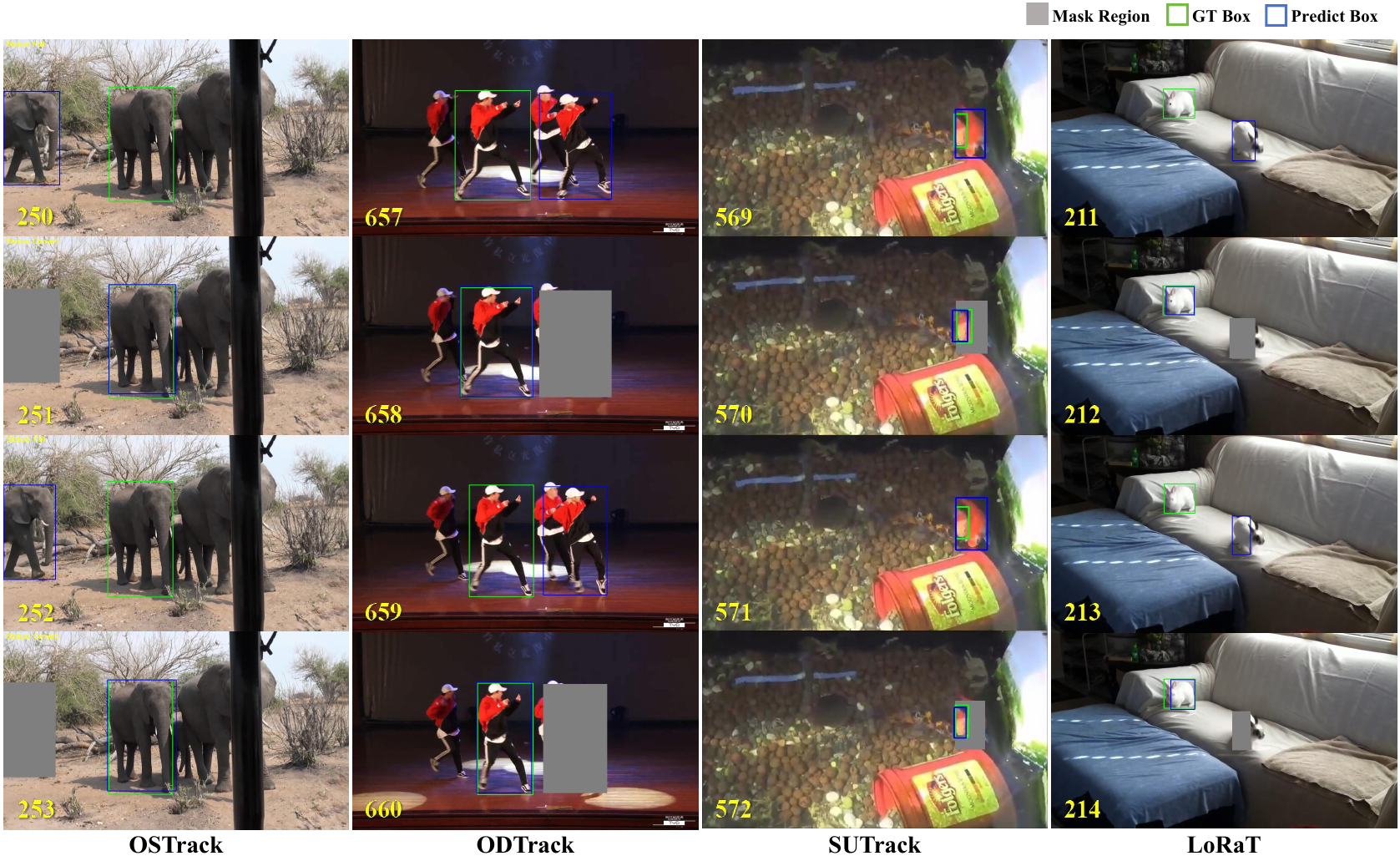}
    \caption{OIM visualization results across different tracking algorithms on LaSOT.}
    \label{fig:OIM_vis_sup}
\end{figure*}

Fig. \ref{fig:OIM_vis_sup} demonstrates the corrective effects of the OIM mechanism across various state-of-the-art tracking algorithms. As shown in the figure, when trackers experience drift (blue prediction boxes deviating from green ground-truth boxes), the OIM mechanism effectively guides all tested trackers back to the correct target by masking interference regions (gray areas) while preserving the ground-truth target region. However, the critical finding is that once the mask guidance is removed, trackers immediately revert to drift behavior, re-tracking similar distractor objects. This phenomenon is consistently observed across different architectural frameworks, including OSTrack \cite{ye2022joint}, ODTrack \cite{zheng2024odtrack}, SUTrack \cite{chen2025sutrack}, and LoRAT \cite{lorat}, providing compelling evidence that existing trackers lack genuine semantic cognitive capabilities and rely solely on shallow visual feature matching for target identification. This ``immediate correction but inability to sustain" behavioral pattern reveals the fundamental limitation of current tracking paradigms: the inability to establish persistent target identity representations, with each frame's decision heavily dependent on current visual input and lacking deep understanding of the target's essential characteristics.

\subsubsection{Representative Failure Cases}

\begin{figure*}[htbp]
    \centering
    \includegraphics[width=1\linewidth]{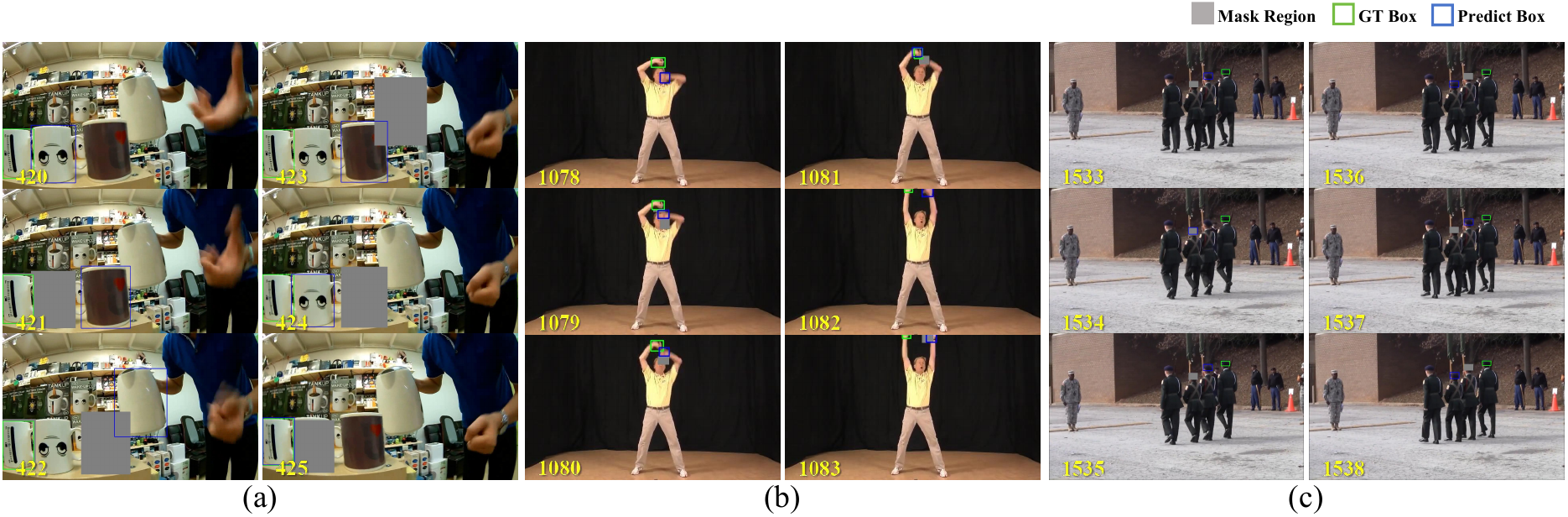}
    \caption{Failure cases of the OIM mechanism revealing fundamental tracker limitations}
    \label{fig:OIM_vis_sup01}
\end{figure*}

Figure \ref{fig:OIM_vis_sup01} presents three representative failure modes of the OIM mechanism that reveal specific technical limitations in current tracking paradigms. Here we use the SOTA algorithm LoRAT \cite{lorat} as an illustrative example.

In case (a), after OIM guidance, the tracker drifts even further from the target. This occurs because existing trackers employ a local tracking paradigm that processes cropped regions around the previous prediction rather than the complete image. When the tracker focuses on interference objects within this limited field of view, it can completely lose sight of the actual target, making the guidance counterproductive.

In case (b), under rapid target motion, the spatially precise masking mechanism fails due to accuracy and latency issues. Since visual guidance operates on results from the previous frame, when objects move at high speeds, the mask cannot perfectly eliminate interference objects in their new positions, causing the mechanism to become ineffective.

In case (c), the tracker continues to predict masked regions even when they appear as uniform gray blocks (125, 125, 125 RGB values). This demonstrates that current trackers lack robust feature modeling capabilities and remain overly dependent on positional priors rather than meaningful visual content, failing to adapt when interference regions are explicitly obscured.

These failure cases illustrate critical technical challenges in current tracking architectures that constrain the effectiveness of external guidance. Together, they underscore the urgent need for semantic-level guidance beyond shallow appearance matching, which directly motivates the SOIBench framework proposed in the main paper.

\begin{algorithm}[!t]
\caption{SOI Challenge Mining Pipeline}
\label{alg:soi_mining}
\begin{algorithmic}[1]
\REQUIRE Trackers $\{\mathcal{T}_1, ..., \mathcal{T}_N\}$; Sequence $\{I_1, ..., I_T\}$; Ground-truth boxes $\{GT_1, ..., GT_T\}$; thresholds $\alpha, \beta, \tau$
\ENSURE Set of SOI frames $S$
\STATE Initialize $S \gets \emptyset$, $t_{\text{last}} \gets -\infty$
\FOR{$t = 1$ to $T$}
    \STATE $\mathcal{C}_t \gets \emptyset$ \hfill \# initialize candidate set
    \FOR{each tracker $\mathcal{T}_i$}
        \STATE $H_t^i \gets \mathcal{T}_i(I_t)$ \hfill \# confidence score map
        \STATE $P_t^i \gets \textsc{MaxPool}(H_t^i, k)$ \hfill \# local maxima pooling
        \STATE $\text{Peaks}_t^i \gets \{(x,y): P_t^i(x,y) \ge \alpha \cdot \max(P_t^i)\}$ \hfill \# peak detection
        \STATE $C_{t,\text{raw}}^i \gets \{GT_t\} \cup \textsc{DecodeBoxes}(H_t^i, \text{Peaks}_t^i)$ \hfill \# raw candidates
        \STATE $C_t^i \gets \textsc{Deduplicate}(C_{t,\text{raw}}^i, \beta)$ \hfill \# IoU-based deduplication
        \STATE $\mathcal{C}_t \gets \mathcal{C}_t \cup C_t^i$
        \STATE \# tracker status determination
        \IF{IoU$(\text{pred}_t^i, GT_t) \ge \tau$ \AND $|C_t^i| = 1$}
            \STATE $V_{t,i} \gets 0$ \hfill \# Correct
        \ELSIF{IoU$(\text{pred}_t^i, GT_t) \ge \tau$ \AND $|C_t^i| > 1$}
            \STATE $V_{t,i} \gets 1$ \hfill \# Compromise
        \ELSIF{IoU$(\text{pred}_t^i, GT_t) < \tau$ \AND $|C_t^i| > 1$}
            \STATE $V_{t,i} \gets 1$ \hfill \# Drift
        \ELSE
            \STATE $V_{t,i} \gets 0$ \hfill \# Fail
        \ENDIF
    \ENDFOR
    \STATE \# multi-tracker collective judgment
    \IF{$\sum_{i=1}^{N} V_{t,i} \ge (\lfloor N/2 \rfloor + 1)$}
        \STATE \# sequence-level optimization
        \IF{Area$(GT_t)/$Area$(I_t) \ge 0.001$ \AND $(t - t_{\text{last}} \ge 30 \lor \text{IoU}(GT_t, GT_{t_{\text{last}}}) < \sigma_1)$}
            \STATE $S \gets S \cup \{t\}$, $t_{\text{last}} \gets t$
        \ENDIF
    \ENDIF
\ENDFOR
\RETURN $S$
\end{algorithmic}
\end{algorithm}

\section{More Details of SOIBench}

This section supplements the details of SOIBench, including the automated SOI challenge mining pipeline and the multi-level semantic annotation protocol.

\subsection{SOI challenge mining pipeline Details}

The SOI frame mining process strictly follows the three-phase design introduced in Sec.~4.1 of the main paper. For completeness, we provide the pseudocode in Algorithm~\ref{alg:soi_mining}, which formalizes candidate extraction, multi-tracker collective judgment, and sequence-level optimization. This pipeline ensures that the selected SOI frames correspond precisely to current trackers’ cognitive boundaries, free from manual bias. The Key hyperparameters are shown in Tab \ref{tab:soibench_hyperparams}.

\begin{table}[!t]
\centering
\small
\renewcommand{\arraystretch}{1.2}
\begin{tabular}{lc}
\toprule
\textbf{Parameter} & \textbf{Value} \\
\midrule
Max-pooling kernel size ($k$) & 5 \\
Candidate threshold ($\alpha$) & 0.6 \\
IoU deduplication threshold ($\beta$) & 0.4 \\
SOI decision IoU threshold ($\tau$) & 0.6 \\
Temporal filtering interval & 30 frames \\
Scene-change IoU threshold ($\sigma_1$) & 0.5 \\
\bottomrule
\end{tabular}
\caption{Hyperparameters for the SOI challenge mining pipeline.}
\label{tab:soibench_hyperparams}
\end{table}

\begin{figure}[!t]
\centering
\begin{tcolorbox}[colback=gray!10,colframe=gray!40,title=Structured Prompt for Initial Generation,width=0.47\textwidth]
\tiny
\textbf{Scene Description:}\\
You are observing an image that contains the tracking target, marked clearly by a green\\
bounding box. There are also several visually similar distractor objects in the scene.\\
The target object is clearly marked.\\

\textbf{Core Warning \& Task:}\\
The green bounding boxes in the image are provided only to help you analyze the scene.\\
Do NOT mention any bounding boxes, coordinates, or technical annotation terms\\
in your description.\\

Your task is to produce a concise, structured, multi-level semantic description of the\\
tracking target, guided strictly by two principles from cognitive linguistics:\\
- \textbf{Concretization} (vivid, specific, and easily imaginable details)\\
- \textbf{Saliency guiding} (highlighting distinctive features that rapidly differentiate the\\
\hspace{0.5em}target from distractors)\\

\textbf{Required Output Format:}\\
Return your answer strictly in this JSON format:\\
\{\\
\hspace{1em}``level1": ``$<$Location Feature$>$",\\
\hspace{1em}``level2": ``$<$Appearance Description$>$",\\
\hspace{1em}``level3": ``$<$Dynamic State Description$>$",\\
\hspace{1em}``level4": ``$<$Distractor Differentiation$>$"\\
\}\\

\textbf{Detailed Level Instructions:}\\
\textbf{- Location Feature}\\
\hspace{0.5em}• Start with a preposition and end with a comma\\
\hspace{0.5em}• Describe semantic location (e.g., ``At the center of the roadway,")\\
\hspace{0.5em}• Never include coordinates or annotation terms\\

\textbf{- Appearance Description}\\
\hspace{0.5em}• Use one of these formats:\\
\hspace{1em}1. ``a/an [adjective(s)] [object]"\\
\hspace{1em}2. ``a/an [adjective(s)] [object] on/in [carrier]"\\
\hspace{1em}3. ``a/an [adjective(s)] [object] held/carried by [carrier]"\\
\hspace{0.5em}• Always include \textbf{color + object type}, plus salient visual features\\

\textbf{- Dynamic State Description}\\
\hspace{0.5em}• Output a complete verb phrase that continues the sentence\\
\hspace{0.5em}• Describe motion/pose/state (e.g., ``is running along the sidewalk")\\

\textbf{- Distractor Differentiation}\\
\hspace{0.5em}• Start phrases with ``to the target's [direction]"\\
\hspace{0.5em}• Autonomously identify elements that may confuse a tracker\\
\hspace{0.5em}• Use clear directional and visual distinctions\\
\hspace{0.5em}• Avoid vague terms like ``different" or ``unlike"\\

Output only the JSON object, without any explanation or markdown syntax.
\end{tcolorbox}
\caption{VLM annotation prompt structure for generating multi-level semantic descriptions.}
\label{fig:vlm_prompt}
\end{figure}

\subsection{Multi-Level Semantic Annotation Process Details}
\label{sec:annotation-details}

\begin{figure*}[htbp]
    \centering
    \begin{subfigure}[b]{0.9\linewidth}
        \includegraphics[width=\linewidth]{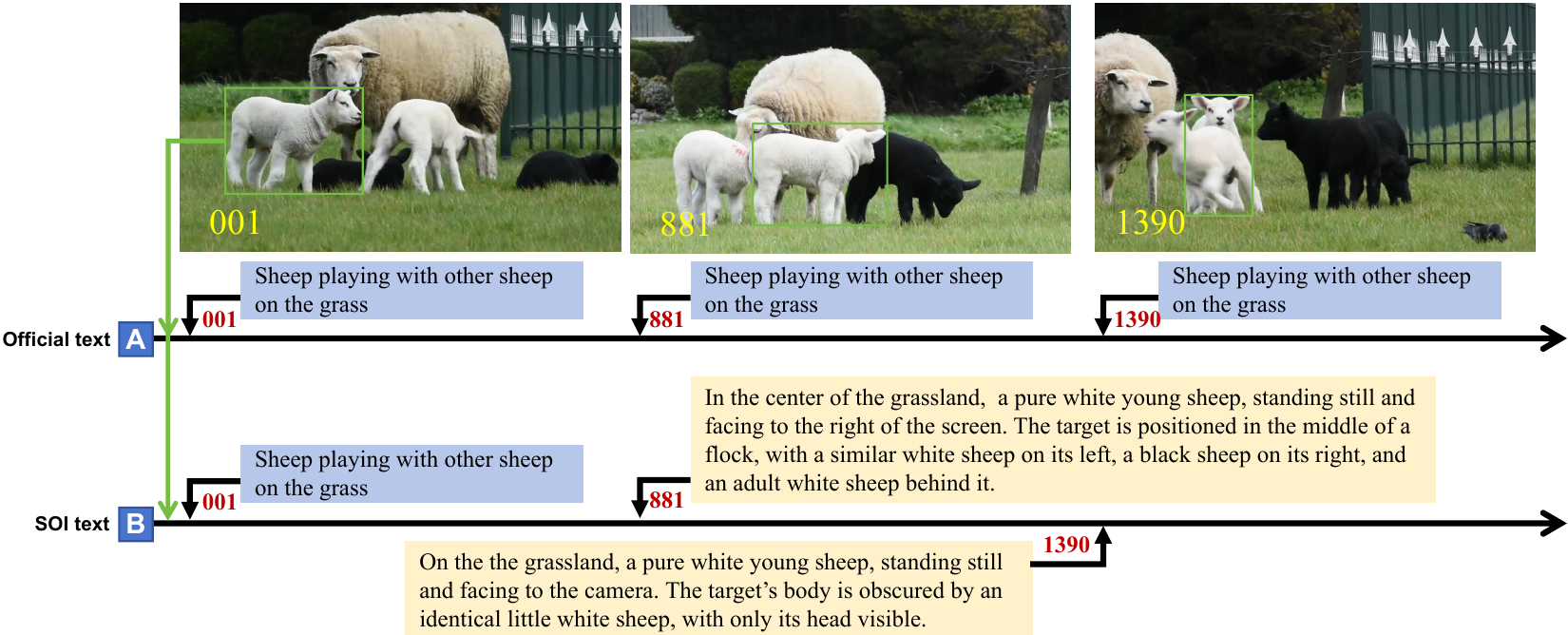}
        \caption{Case 1} 
    \end{subfigure}

    \vspace{0.5em}
    \begin{subfigure}[b]{0.9\linewidth}
        \includegraphics[width=\linewidth]{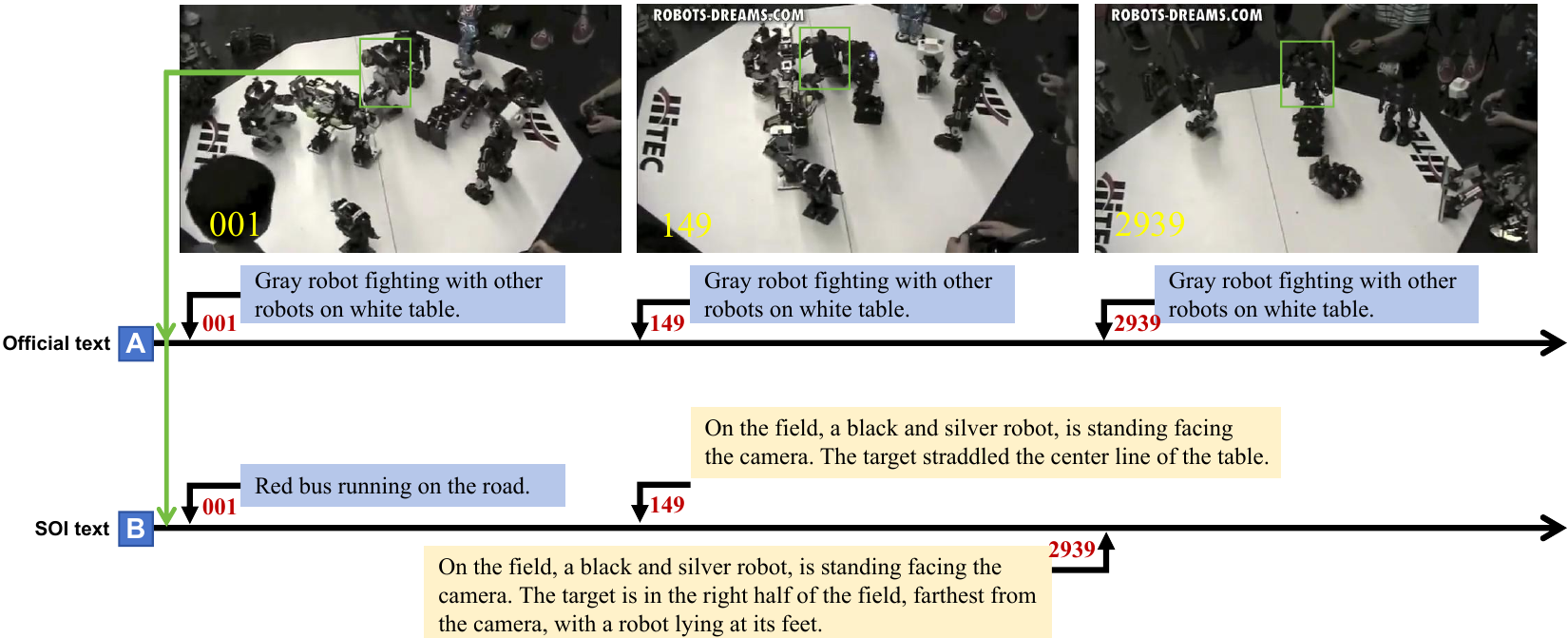}
        \caption{Case 2} 
    \end{subfigure}

    \vspace{0.5em}
    \begin{subfigure}[b]{0.9\linewidth}
        \includegraphics[width=\linewidth]{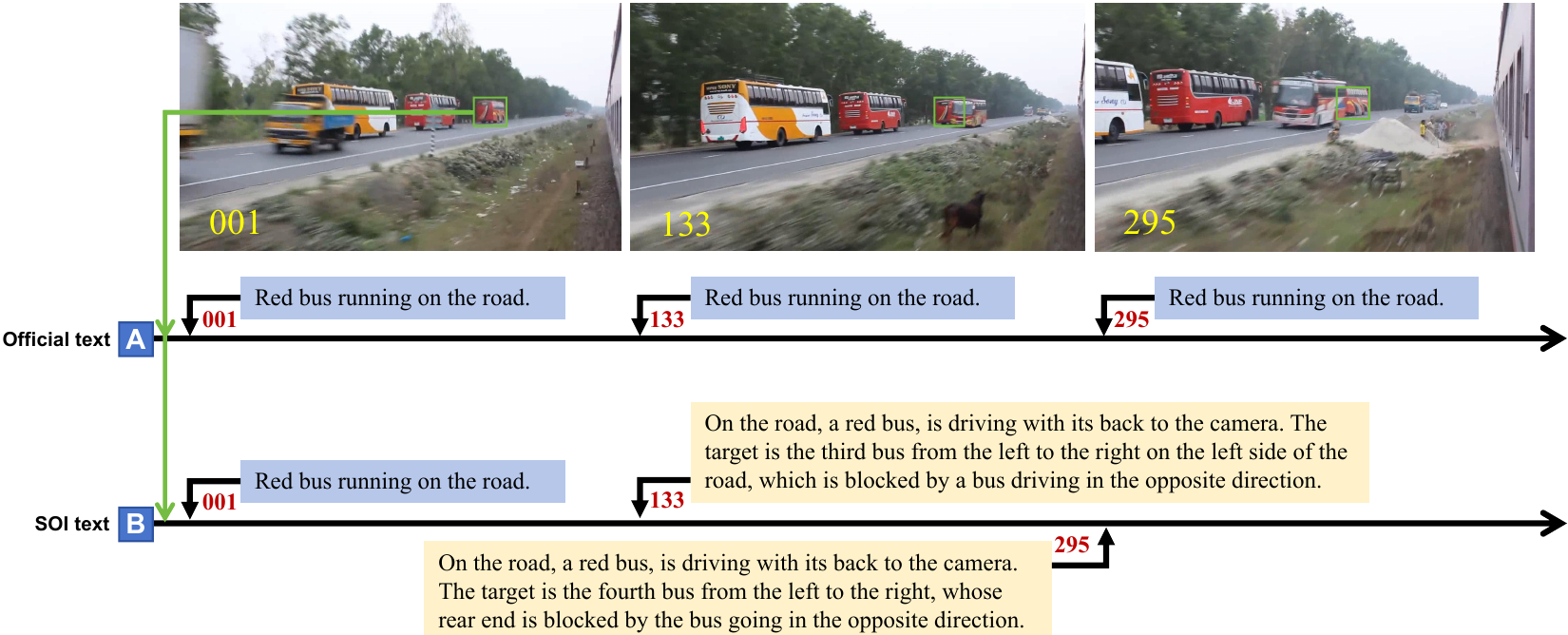}
        \caption{Case 3} 
    \end{subfigure}

    \caption{Instantiated cases from LaSOT$_{\text{SOI}}$. (a)--(c) illustrate representative frames with SOI-enhanced annotations.}
    \label{fig:lasot_soi_cases}
\end{figure*}

\subsubsection{Annotation Framework Design}

Our semantic annotation framework is grounded in cognitive linguistics and tailored to provide hierarchical textual guidance for SOI challenges in visual tracking. It follows two key principles:

\begin{itemize}
    \item \textit{Concretization}: Descriptions should be vivid, specific, and easily imaginable, offering concrete visual anchors rather than abstract labels.
    \item \textit{Saliency Guiding}: Descriptions should emphasize distinctive features that enable rapid differentiation of the target from visually similar distractors.
\end{itemize}

\subsubsection{Hybrid Annotation Workflow}
We adopt a hybrid annotation workflow that combines the efficiency of large-scale VLMs with the precision of human experts:

\begin{itemize}
    \item \textit{VLM-Based Initial Generation}: Qwen2.5-VL-32B \cite{wang2024qwen2} serves as the primary annotation engine. Each mined SOI frame is processed with structured prompts designed according to our cognitive linguistics principles. The model outputs initial four-level descriptions in a JSON format, as illustrated in Figure~\ref{fig:vlm_prompt}.
    \item \textit{Human Review and Refinement}: A trained annotation team systematically reviews and refines the VLM-generated drafts, focusing on:
    \begin{itemize}
        \item Hallucination Mitigation: Identifying and correcting hallucinated details or misinterpreted scene content.
        \item Semantic Clarity Enhancement: Strengthening descriptions that lack sufficient discriminative power or contain ambiguous expressions.
    \end{itemize}
\end{itemize}

Quality assurance is enforced through multiple rounds of consistency checks, ensuring strict adherence to both cognitive linguistics principles and annotation format specifications. This hybrid workflow leverages the scalability of VLMs while guaranteeing the semantic precision required for reliable SOI guidance.










\section{More Details of the Construction of LaSOT$_{\text{SOI}}$} 

Due to space constraints in the main manuscript, we provide additional details on the construct of LaSOT$_{\text{SOI}}$.

We employ four state-of-the-art trackers as consensus judges for automatic SOI frame identification: OSTrack \cite{ye2022joint}, ODTrack \cite{zheng2024odtrack}, SUTrack \cite{chen2025sutrack}, and LoRAT \cite{lorat}. These trackers represent the current cognitive upper limits of tracking algorithms, ensuring that the mined SOI frames possess both representativeness and comprehensiveness across various challenging scenarios. 


In addition, we present visualized examples of the SOI annotations in Fig.~\ref{fig:lasot_soi_cases}. 
These cases demonstrate how the instantiated SOI-enhanced descriptions go beyond the generic official annotations 
by explicitly highlighting distinctive spatial cues, occlusion details, and distractor differentiation. 
Such fine-grained visual-textual alignment provides more reliable semantic guidance, 
facilitating robust target identification under visually confusing conditions.

\section{VLM Grounding Experiment Details.} 
We follow the official evaluation guidelines of the respective VLMs to conduct our grounding experiments. The goal is to test whether VLMs can accurately localize targets in SOI frames using our structured semantic descriptions.  

\textbf{Models Evaluated.} 
We benchmark four representative VLMs: Qwen2.5-VL-3B, Qwen2.5-VL-7B\cite{wang2024qwen2}, and DeepSeek-VL2\cite{wu2024deepseekvl2}. 
Human annotations are included as the upper bound for comparison.

\textbf{Human Grounding Experiment.} 
To establish a reliable upper bound, we conducted a human grounding study involving three researchers with prior experience in visual tracking and annotation. 
All participants underwent a brief training session to familiarize themselves with the SOI annotation protocol. 
During the study, each participant was presented with the same structured semantic descriptions used in the VLM experiments, 
and was required to localize the target in the current frame by drawing a bounding box.  
The dataset was evenly divided among the participants, with approximately one-third of the SOI frames assigned to each. 
Partial overlap between subsets was introduced to assess inter-annotator agreement.

\textbf{VLMs Implementation.} 
Experiments were conducted with both API-based inference (DeepSeek-VL2) and local inference (Qwen2.5-VL using Transformers).  
All code is implemented in Python and aligned with the official VLM testing specifications. 
Table~\ref{tab:vlm_prompts} summarizes the prompts used for different models, all following the optimal implementations recommended by the official guidelines. 
We set the temperature to 0.1 for stable outputs, with a maximum generation length of 2048 tokens.

\begin{table}[h]
\centering
\small
\renewcommand{\arraystretch}{1.3}
\begin{tabular}{p{0.95\linewidth}}
\toprule
\textbf{Prompt for Qwen-VL \cite{wang2024qwen2}} \\
\hline
Please identify and locate the target object based on the description:  
``\textit{[SOI description]}''.  
Output its bounding box coordinates using JSON format. \\
\midrule

\textbf{Prompt for DeepSeek-VL2 \cite{wu2024deepseekvl2}} \\
\hline
\texttt{<image>}  
\texttt{<|ref|>} \textit{[SOI description]} \texttt{<|/ref|>}. \\
\bottomrule
\end{tabular}
\caption{Prompts used for different VLMs in the SOI grounding experiments.}
\label{tab:vlm_prompts}
\end{table}

\bibliography{aaai2026}

\end{document}